%% file: main.tex
\documentclass[10pt,twocolumn,letterpaper]{article}

\usepackage{cvpr}              % To produce the CAMERA-READY version
\input{preamble}

\definecolor{cvprblue}{rgb}{0.21,0.49,0.74}
\usepackage[pagebackref,breaklinks,colorlinks,citecolor=cvprblue]{hyperref}
\usepackage{graphicx}% http://ctan.org/pkg/graphicx

\def\paperID{*****} % *** Enter the Paper ID here
\def\confName{CVPR}
\def\confYear{2024}

\title{Embedding Compression for Efficient Re-Identification}

\author{Luke McDermott\\
Modern Intelligence, UC San Diego\\
{\tt\small luke@modernintelligence.ai}
}

\begin{document}
\maketitle

\begin{abstract}
Real world re-identfication (ReID) algorithms aim to map new observations of an object to previously recorded instances. These systems are often constrained by quantity and size of the stored embeddings. To combat this scaling problem, we attempt to shrink the size of these vectors by using a variety of compression techniques. In this paper, we benchmark quantization-aware-training along with three different dimension reduction methods: iterative structured pruning, slicing the embeddings at initialize, and using low rank embeddings. We find that ReID embeddings can be compressed by up to 96x with minimal drop in performance. This implies that modern re-identification paradigms do not fully leverage the high dimensional latent space, opening up further research to increase the capabilities of these systems.
\end{abstract}

\section{Introduction}
Object re-identification (ReID) is an image retrieval task that requires mapping novel views of an object to previous instances. ReID models are typically trained with contrastive methods to encode different views of the \textit{same} object to a similar area in the latent space, while simultaneously pushing views of \textit{different} objects further away in the latent space. At inference time, incoming embeddings are matched to the previous instance with the smallest euclidean distance. 
Common machine learning pipelines for this task involve storing embeddings of each object instance for future re-identification. Such pipelines are not scalable as the number of unique objects increases. In constrained deployment scenarios, like offline edge devices, compression is needed to alleviate storage issues. While the original model’s weights can also be compressed, this work primarily focus on reducing the cost of storing embeddings as this is undocumented in the scope of ReID.

We find that ReID embeddings do not fully span the latent space. These can be compressed by up to 96x with only a 4\% drop in accuracy on Market-1501 \cite{Market}. This work studies the tradeoffs between different dimension reduction methods such as using a fixed subset of dimensions (slicing), using a learnable low rank factorization, and using iterative compression with structured pruning \cite{pruning}. Across these methods, we also perform quantization-aware training \cite{quantization} for a further 4x compression. 

These results, while boasting impressive compression, uncover a much larger problem in this field's models: the high-dimensional latent space is severely underutilized. More research is needed to determine if these models should use smaller embeddings as a default, or if they should attempt to fill out the large latent space with additional regularization.

\section{Background}
\begin{figure}
  \includegraphics[width=3.5in]{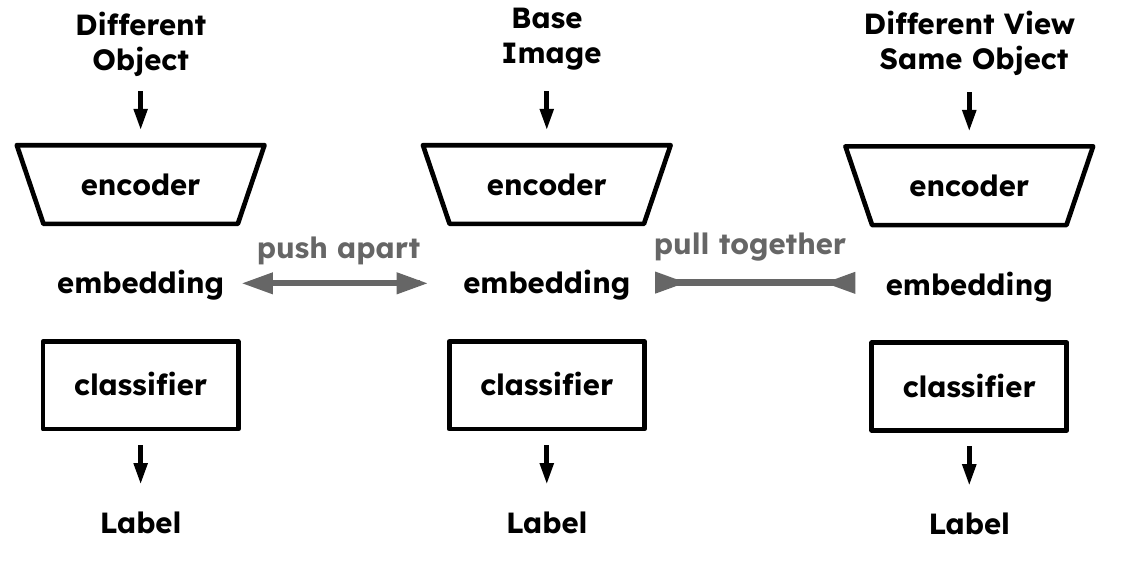}
  \caption{Re-identification Training Paradigm}
  \label{fig:reid}
\end{figure}

In the typical re-identification training paradigm, some encoder, either a CNN or ViT, is trained to output robust and informative features. These features are trained such that the same object will be mapped to a similar area in the latent space, agnostic to the view. Here, it is helpful to consider “different views of the same object” as an extreme form of data augmentation. As shown in Figure \ref{fig:reid}, these features are trained with triplet loss, where we sample an “anchor” image along with a “positive” and “negative” sample. The positive sample has the same ID or label as the anchor, while the negative has a different ID. Each sample passes through the same shared encoder and we apply triplet loss on these embeddings. To prevent collapse during training time, we also use a linear classifier to predict which ID these objects belong to. This maintains that these embeddings are informative. At inference time, the linear classifier can be discarded as “re-identifications” will only be decided from the embeddings alone. 

During deployment, each embedding must be saved so that future samples can be re-identified. Then, after sufficient amount of samples, incoming query embeddings will be matched to a gallery of historical vectors. If there is a close enough match (measured by euclidean distance), then a prediction is made.

There is a significant amount of work that focuses on improving the capability of these systems, most notably \textit{TransReID} \cite{TransReID} and \textit{Bag of Tricks} \cite{BagofTricks}. ReID models can be applied to a broad range of objects such as people \cite{SOLIDER}, vehicles \cite{VehicleReID}, or even objects observed in multiple modalities \cite{UniCat}. While most of the literature is application focused, there is a great need for better understanding the training dynamics of these models.

This training paradigm draws very similar ideas to self-supervised learning (SSL) across different views. Shwartz-Ziv et al.'s work, ``To Compress or Not Compress" \cite{ToCompress}, discusses how these SSL methods can draw from first principles in information theory. According to the Deep Information Bottleneck \cite{bottleneck}, more compressed features can improve generalization as they reduce the potential for noisy information to exist. However, compression comes at the cost of underfitting and creating uninformative features. This balance is especially crucial in noisy, small datasets which are all too common in re-identification benchmarks. According to \citeauthor{ToCompress}, our task and datasets holds under the MultiView Assumption: the shared information between two views holds task-relevant information. Thus, we should be minimizing the amount of mutual information between the input and its associated features while maximizing the mutual information between features from different views of the same object. In the context of this paper, ``minimizing the mutual information between input and the features" roughly equates to how compressed our embedding is. 

%Previous work in this field have made advances strictly on smaller datasets, which have yet been proven at scale. This work provides other researchers the tools to use datasets with more unique IDs. We also hope to motivate work regularization methods for re-identification to ensure all dimensions are adequately used.

\section{Methods}
\begin{figure}
  \includegraphics[width=3.2in]{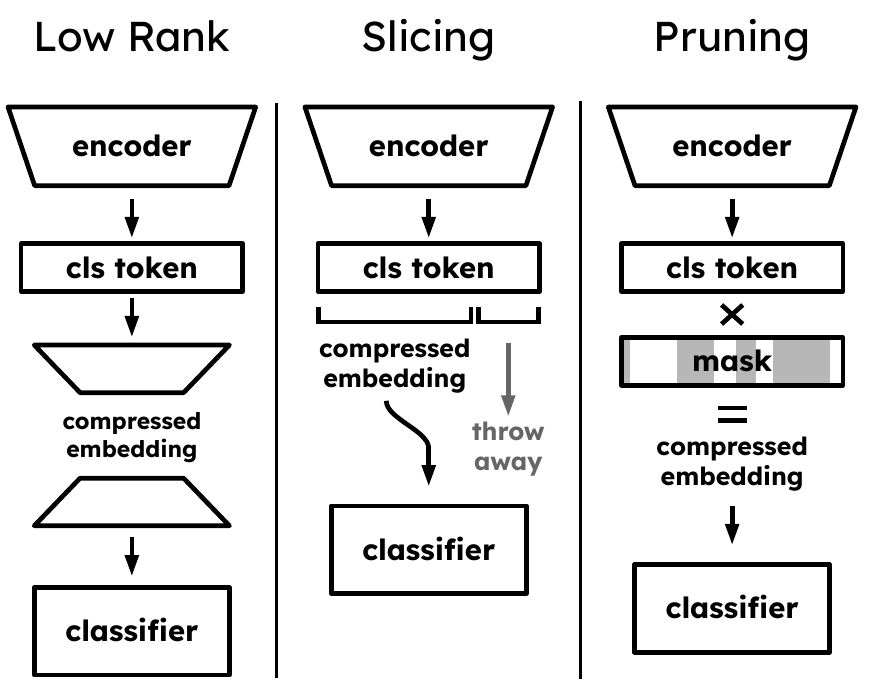}
  \caption{Dimension Reduction Methods}
  \label{fig:method}
\end{figure}

We benchmark common compression techniques such as structured pruning \cite{pruning}, quantization \cite{quantization}, low rank factorization, and slicing the embedding at initialization. Each method is illustrated in Figure \ref{fig:method}.

\paragraph{Slicing.}
To measure other methods against a strong baseline, we train the models with only a subset of dimensions from initialization. In our implementation, this is carried out by slicing the cls token. While being a naive approach, we use this baseline to determine if other complex methods are truly needed, as they require either increased parameterization or training time.

\paragraph{Low Rank Embeddings.}
Since the performance of our embeddings are measured by its euclidean distance from others, we cannot use off-the-shelf factorization methods like PCA. Instead, we opt for a learnable factorization by augmenting the model with a low rank matrix placed after the cls token. Triplet loss is applied to this compressed embedding, inside the low-rank matrix. We still expand these features and connect them to the classifier to prevent collapse. 

\paragraph{Iterative Structured Pruning.}
Iterative pruning \cite{pruning} can be thought of as a continual compression over training and is a more-informed version of slicing since we use post-training information to select unimportant dimensions. To carry out this procedure, we first train our model until convergence, then remove the least important dimensions. We continue the process by retraining the model, removing even more dimensions, and iterating until our compression target is reached. We value the least significant dimensions as those with the smallest frobenius norm on the training set. Due to the iterative nature, this compression algorithm requires the largest computational cost at training time, as each retraining uses 20\% of the epochs used during normal training. We chose 20\% as this equates to doubling our total training time for the largest compression ratio that we test on.

\paragraph{Quantization.}
Orthogonal to any other compression methods stated above, the number of bits used to represent our embedding can be reduced with quantization \cite{quantization}. Rather than storing tensors in float-32, they are reduced to int-8 with quantization-aware training. This means the embeddings are quantized on the forward pass, but have full precision gradients on the backward pass. For our initial experiments in this short paper, we use vanilla uniform quantization.

\section{Results}
In our experiments, each of the dimension-reduction methods are recorded across various compression rates, as well as, its performance when paired with quantization-aware training. We measure these methods across three different re-identification datasets with a standard ViT-Base backbone \cite{ViT}. Each method is evaluated across 6 different compressed size with 768 as our default embedding dimension: 576, 480, 384, 192, 64, \& 32 dimensions per embedding. We use the same baseline unimodal architecture as described in \cite{UniCat}.

\subsection{Market-1501}
Market-1501 \cite{Market} is a popular benchmark for person re-identification, consisting of 1501 different individuals with 32,668 total image samples. This dataset was collected by 6 different stationary cameras.

\begin{figure}[h]
  \includegraphics[width=3.25in]{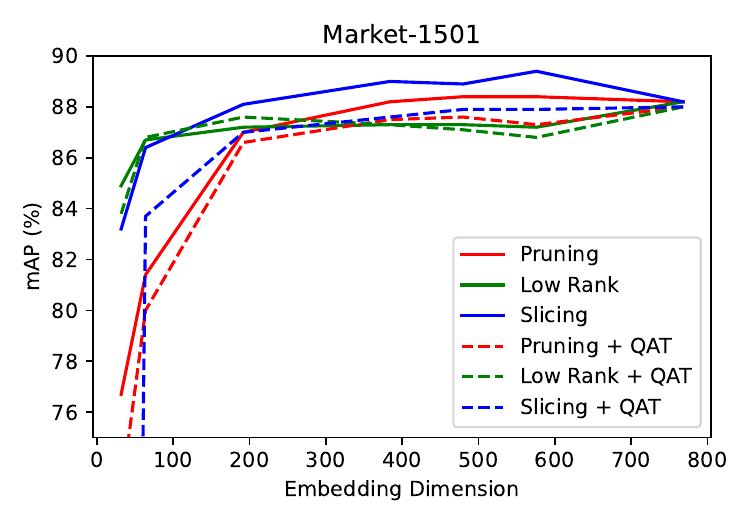}
  \caption{Performance of each compression method across different compression ratios on Market-1501. The original embedding dimension is 768. Higher mAP is better performance. Smaller embedding dimension is more efficient.}
  \label{fig:market}
\end{figure}

As shown in Figure \ref{fig:market}, training from initialization with only a subset of dimensions, notated as slicing, sees an increased performance over the full dimensional model by using a slight compression. This constraint may be reducing overfitting to the view-specific information in the training set. Using a low rank embedding does not see any improvement;  however, it does provide the best performance with few dimensions.

\subsection{PRW}
PRW \cite{PRW} is a different processing and labeling of the same raw footage as Market-1501. The main difference is that video frames are annotated allowing for an order of magnitude more gallery images. Since this dataset is a different annotation of Market, IDs in one do not correspond to the other. There are 11,816 total frames annotated, making up the 34,304 samples. Instead of the 1501 IDs in Market-1501, there are only 932 labels here. This dataset provides a more realistic benchmark for deploying ReID in real world settings due to the large gallery from video frames. This should be a harder task, especially as our embedding space is compressed.
\begin{figure}[h]
  \includegraphics[width=3.25in]{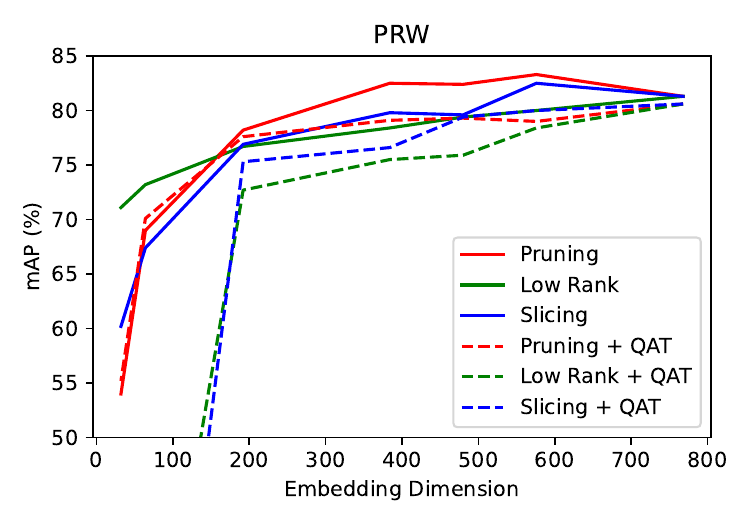}
  \caption{Performance of each compression method across different compression ratios on PRW.}
  \label{fig:prw}
\end{figure}

Compared to the results on Market, Figure \ref{fig:prw} shows that pruning provides the best performance at low compression ratios\footnote{Compression ratio is defined as the original embedding size divided by the compressed size.}. In our trials, we found that PRW converged faster during retraining compared to Market or PRAI. This can either be due to PRW generally training better or that the pruning criteria, lowest frobenius norm, aligns better with this datasets distribution. We continue to see the trend of the low rank method performing best on the smallest embedding dimension; however, this does not hold under quantization.

\subsection{PRAI}
\begin{figure}[h]
  \includegraphics[width=3.25in]{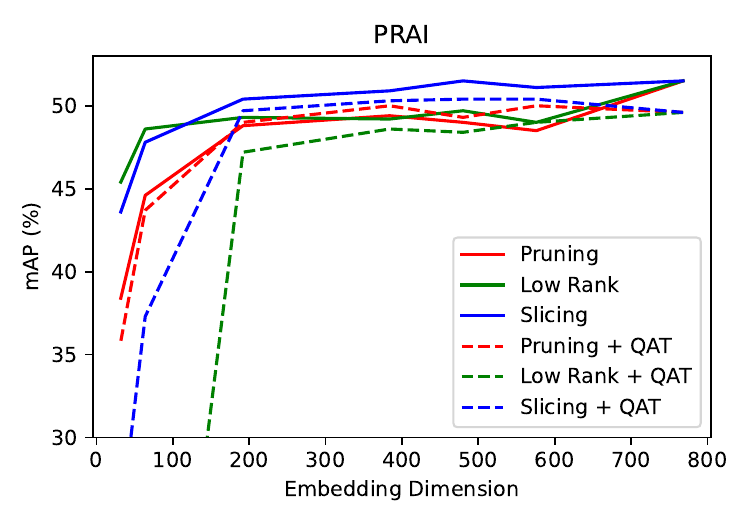}
  \caption{Performance of each compression method across different compression ratios on PRAI.}
  \label{fig:prai}
\end{figure}

PRAI \cite{PRAI} is an aerial imagery dataset for person re-identificaton. There are 39,461 total images of 1581 individuals taken from UAVs. Aerial footage introduces a unique challenge as samples are further away, taken at an angle, and have more diverse environments compared to a stationary camera. 

In Figure \ref{fig:prai}, we see a similar trend to Market; however, quantization drastically underperforms here compared to the dimension reduction methods. Due to the aerial view of these images, people are represented with much less information, likely translating to less diversity of relevant features. One possible reason for these results could be that the few features are strongly used, implying that less dimensions are needed to represent them without superposition \cite{superposition}. However, these few features require high precision to differentiate between the labels in the dataset.

\section{Discussion}
As seen in our results, the choice of compression is a function of the target compression ratio and the dataset. Slicing the cls token tends works the best at lower compression rates, with the exception of pruning on PRW. The low rank embedding provides the best performance in extremely constrained settings for all datasets. 

Slicing is a subset of low rank factorization \footnote{For some decomposed matrices A \& B of shape $(d,k)$ and $(k,d)$ respectively, such that $d$ is the original embedding dimension and $k$ is the compressed size. Let $A[:k,:]$ be the identity matrix, and $A[k:,:]$ be the zero matrix. Similarly, form $B$ to follow this pattern.} , implying that either the low rank method trains to less generalizable solutions in unconstrained settings, or it cannot optimize over the larger space. Likely, the former is true. According to the deep information bottleneck \cite{bottleneck}, the generalization issue can be alleviated with heavier compression at the cost of underfitting. This view may explain why factorization only performs better when decomposing to a small rank. At high ranks, the matrix overfits to the training data.

Using iterative pruning in our setting did not show any improvement on Market \& PRAI, likely due to how the retraining was structured. On the PRW dataset, we observed a much faster convergence during retraining. As a reminder, the models only retrained for 20\% of the total training epochs to reduce the amount of overhead of using this method. Any more would seem infeasible to use in real world applications as compressing to 64 dimensions already doubles the total training time. Considering the limitations of this study, we can only reasonably conclude that pruning does not outperform slicing or low rank decomposition only in scenarios with a limited retraining budget. With a much larger training budget or datasets with faster convergence, we still hypothesize that pruning should at least match the slicing baseline. Regardless of this performance on PRW, pruning does not seem to be worth doubling the training costs in most scenarios.

Across all methods, we see that a large embedding space is not needed for accurate re-identification using our current training paradigms. The field now needs to reconsider whether additional regularization is needed to make use of the full dimensional space. Non-contrastive self-supervised learning methods may be the answer here for better utilization of our latent space. From an information-theoretic perspective, the mutual information between the embedding and our label is not appropriately maximized \cite{ToCompress}. This results in a large amount of view-specific noise existing in these embeddings.

Until better paradigms are created, we urge researchers and practioners alike to consider compressing the final embedding for faster re-identification and less resource usage.

In the future, we plan to provide a more robust benchmark of methods, using more methods, datasets, and tasks. Quantization seems to be the lowest hanging fruit as this work can incorporate more cutting edge methods like SmoothQuant \cite{SmoothQuant} or outlier suppression with QuIP \cite{QuIP}. In addition, there is a large similarity between object re-identification in computer vision and text retrieval or RAGs (retrieval augmented generation) in language. We plan to benchmark these compression methods across these domains, lowering the cost of storing embeddings for everyone.

%limitations

{
    \small
    \bibliographystyle{ieeenat_fullname}
    \bibliography{main}
}


\end{document}

%% file: preamble.tex
\usepackage[dvipsnames]{xcolor}